\documentclass[sigconf=false, nonacm=true, review=false, anonymous = false,screen=true]{acmart}

\usepackage{mathtools}
\usepackage{algorithm}
\usepackage{algpseudocode}
\usepackage{xspace}
\usepackage{soul}
\usepackage{amsfonts}
\usepackage{amsmath}
\newtheorem{definition}{Definition}[section]
\newtheorem{assumption}{Assumption}

\usepackage{mathrsfs}
\usepackage{subcaption}
\usepackage{mathtools}

\usepackage{multirow}

\graphicspath{{pic/}}

\usepackage{hyperref}

\allowdisplaybreaks

\newcommand{\ignore}[1]{}

\newcommand{\update}[1]{#1}

\def\HiLi{\leavevmode\rlap{\hbox to .92\linewidth{\color{yellow!50}\leaders\hrule height .8\baselineskip depth .5ex\hfill}}}
\def\HiLii{\leavevmode\rlap{\hbox to .88\linewidth{\color{yellow!50}\leaders\hrule height .8\baselineskip depth .5ex\hfill}}}

\newcommand{\R}{\mathbb{R}}
\newcommand{\Z}{\mathbb{Z}}

\newcommand{\E}{\ensuremath{\operatorname{E}}} 

\renewcommand{\epsilon}{\varepsilon}
\newcommand{\eps}{\varepsilon}

\renewcommand{\Pr}{\mathbb{P}}

\DeclareMathOperator*{\argmin}{arg\,min}
\DeclareMathOperator*{\argmax}{arg\,max}

\newcommand{\BF}[1]{
	\relax
	\ifmmode
	\ifcat\noexpand#1\relax 
		\boldsymbol{#1}     
	\else
		\mathbf{#1}
	\fi
	\else
		\textbf{#1}
	\fi
}

\newcommand{\OneMax}{\textsc{OneMax}\xspace}

\newcommand{\Rugg}{\textsc{Ruggedness}\xspace}

\newcommand{\bbobmixint}{COCO/BBOB-MIXINT\xspace}

\begin{document}




\title{Representation-agnostic distance-driven perturbation for optimizing ill-conditioned problems}


\begin{abstract}
    Locality is a crucial property for efficiently optimising black-box problems with randomized search heuristics.
    However, in practical applications, it is not likely to always find such a genotype encoding of candidate solutions that this property is upheld with respect to the Hamming distance.
    At the same time, it may be possible to use domain-specific knowledge to define a metric with locality property.
    We propose two mutation operators to solve such optimization problems more efficiently using the metric.
    The first operator assumes prior knowledge about the distance, the second operator uses the distance as a black box.
    Those operators apply an estimation of distribution algorithm to find the best mutant according to the defined in the paper function, which employs the given distance.
    For pseudo-boolean and integer optimization problems, we experimentally show that both mutation operators speed up the search on most of the functions when applied in considered evolutionary algorithms and random local search. 
    Moreover, those operators can be applied in any randomized search heuristic which uses perturbations.
    However, our mutation operators increase wall-clock time and so are helpful in practice when distance is (much) cheaper to compute than the real objective function.
\end{abstract}


\author{Kirill Antonov}
\affiliation{%
    \institution{LIACS, Leiden University}
    \streetaddress{Niels Bohrweg 1}
    \city{Leiden}
    \postcode{2333}
    \country{The Netherlands}}
\email{
k.antonov@liacs.leidenuniv.nl}
\orcid{0000-0002-8757-8598}
\author{Anna V. Kononova}
\affiliation{%
    \institution{LIACS, Leiden University}
    \streetaddress{Niels Bohrweg 1}
    \city{Leiden}
    \postcode{2333}
    \country{The Netherlands}}
\email{a.kononova@liacs.leidenuniv.nl}
\orcid{0000-0002-4138-7024}
\author{Thomas B{\"a}ck}
\affiliation{%
      \institution{LIACS, Leiden University}
      \streetaddress{Niels Bohrweg 1}
      \city{Leiden}
      \postcode{2333}
      \country{The Netherlands}}
\email{t.h.w.baeck@liacs.leidenuniv.nl}
\orcid{0000-0001-6768-1478}
\author{Niki van Stein}
\affiliation{%
    \institution{LIACS, Leiden University}
    \streetaddress{Niels Bohrweg 1}
    \city{Leiden}
    \postcode{2333}
    \country{The Netherlands}}
\email{n.van.stein@liacs.leidenuniv.nl}
\orcid{0000-0002-0013-7969}




\maketitle

\section{Introduction}

Randomized search heuristics (RSH) are convenient off-the-shelf solvers of difficult optimization problems \cite{auger2011theory, rozenberg2012handbook}. While they do not necessarily guarantee to locate the exact solution to any problem, they aim to approximate a solution of sufficient quality in a feasible time frame. Applications of such algorithms require the user to define a search space $X$, where possible solutions are located, and the objective function $f$ with domain $X$. This function defines the considered problem and usually is expensive to evaluate (in terms of computation cost, licenses or resources). In the scope of the current work, we consider finite set $X \subseteq \Z^n$, function $f$ with image $\R$ and algorithms that aim to \textit{maximize} function $f$, meaning finding such $x^* \in X$ that \[\max\{f(x) \mid x \in X\} - f(x^*) \approx 0\]

Evolutionary algorithms (EAs) are RSH inspired by the process of natural evolution. Such algorithms, supported by theory \cite{jansen2013analyzing, doerr2019theory, lengler2020drift}, show promising applications to pseudo-boolean/integer problems \cite{li2013mixed, doerr2019benchmarking, dushatskiy2021parameterless} including real-world scenarios \cite{back1995evolution, grossmann2002special, slowik2020evolutionary}. One of the main classical assumptions and the most important requirement to the problem optimized with EAs is strong causality \cite{rechenberg1989evolution} also known as contiguity and locality \cite{raidl2005empirical}. It means that a small change in the candidate solution leads to a solution with similar performance according to the fitness function. The significance of the change is measured according to the distance metric $d: X^2 \to \R$ \cite{jansen2013evolutionary}, for example, Hamming distance in the case of pseudo boolean domains, or euclidean distance in the case of $\R^n$ domains. We will say that metric $d$ is \textit{strong causal} (SC) for the set of problems $\mathcal{F}$ if all the problems in the set have locality property with regard to metric $d$. It is known that the existence of such SC metric for $\mathcal{F}$ guarantees that $\mathcal{F}$ is not closed under permutations of the search space $X$ \cite{jansen2013general}. In this case, the NFL theorem \cite{schumacher2001no} guarantees the existence of an algorithm that outperforms other algorithms on average for functions in $\mathcal{F}$. So the existence of SC metric for $\mathcal{F}$ is sufficient to design a problem solver for $\mathcal{F}$ that performs above average.

The SC metric may exist for a set of functions that a practitioner solves, but obtaining this metric usually requires a certain level of expertise in the domain. It means that the SC metric gives away some information on the structure of the considered problems. While EAs do not require any information on the domain of the problem, it is well known that problem-specific knowledge significantly increases the performance of EAs \cite{droste2003design}. We identify 4 main categories of approaches to take advantage of different kinds of domain-specific information in EA:

\textit{1) Asymmetric mutation operator.} Methods in this category use the information about the problem to modify the mutation operator employed by an EA. Doerr et al \cite{doerr2006speeding, doerr2007speeding} consider the Eulerian cycle problem and rigorously show that asymmetry in mutation operator, for example avoiding certain modifications, leads to a performance increase compared to EA with classical mutations. Raidl et al~\cite{raidl2006biased} show the benefits of EA with a mutation operator that chooses rather edges with smaller weights to solve the minimum spanning tree problem and the traveling salesmen problem. Jansen and Sudholt analyzed the influence of a bias in the mutation operator when it is applied to pseudo boolean functions \cite{jansen2010analysis}. They considered a mutation operator that flips 1 with a higher probability when a solution contains more ones and vice-versa. The work rigorously shows that such bias speeds up $(1+1)$-EA equipped with such operator on three of the functions they considered, and slows down the algorithm on another function. Their work was extended by proposing an adaptation of the degree of asymmetry during the optimization process \cite{rajabi2020evolutionary}.

\textit{2) Enhanced representation.} Traditionally, there are no specific restrictions on the used representation, but there is a number of recommendations \cite{rothlauf2006representations}. In the EA community, the mainstream is to simply use the most natural representation of the individual \cite{rudolph2012handbook}. However, finding such a representation may already require expert-level knowledge in the domain of the problem. Many works with a focus on practical applications prove this, for example, Yu et al applied an EA to the object decomposition for 3D printing \cite{yu2017evolutionary}, where they used a Binary Space Partitioning tree encoded as an array of continuous numbers. The design of such representation requires an understanding of the principles of object decomposition. In the area of spatial design creation, the challenging problem of defining a search space that does not significantly restrict the space of all possible solutions was subjected to comprehensive research in \cite{pereverdieva2023prism}. Moreover, standard EA may work significantly slower when the representation is not carefully chosen. Doerr et al developed such a representation for the Eulerian cycle problem that allowed to develop an EA with the arguably best reachable by RSH expected run-time \cite{doerr2007adjacency}.

\textit{3) Data-driven approaches.} This category covers automated algorithm selection \cite{kerschke2019automated, kostovska2022per, ye2021leveraging, long2022learning, guo2019new} and dynamic algorithm configuration \cite{biedenkapp2020dynamic, adriaensen2022automated}, where automatically extracted features of the problem are used to suggest an optimization method or a configuration of the algorithm.

\textit{4) Diverse set of candidate solutions.} Diversity measure of the set of candidate solutions can be evaluated using a domain-specific distance, as it is suggested in \cite{solow1994measuring}. It is known, that more diverse sets of parents in EA facilitate global exploration properties of the algorithm, reduce the risk to get stuck in local optima and allow obtaining multiple dissimilar high-quality solutions \cite{sudholt2020benefits}. Niching methods were developed to maintain diverse sets of individuals in EA, for a comprehensive overview see \cite{das2011real, preuss15, preuss2021multimodal}. Distance-based niching methods have a great number of successful applications, which shows the advantage of distance knowledge. For example, Dynamic Peak Identification niching \cite{shir2009niching} maintains high-performing solutions with a pairwise distance larger than the given lower bound. This method was successfully applied to find a number of high-performing designs of lenses in a single run of an EA in \cite{kononova2021addressing}. In the domain of airfoil optimization, a variation of niching was proposed, where the algorithm alternates between pure quality and distance-based diversity optimizations which allows to obtain better-performing solutions \cite{reehuis2013novelty, reehuis2013guiding}. 

As we see from the examples of approaches in categories 1 and 2, the user needs to have deep knowledge of both: the domain of the problem, and the optimization method. Approaches in category 3 have the disadvantage that the predicted algorithms/configurations do not always perform well, especially when a new (unseen) problem is dissimilar from the problems that the approaches are trained for. Approaches in category 4 leave the hard work to pick a distance that represents the problem well to the expert in the domain and allows using this distance as a black-box in the optimization algorithm. However, this distance has influence only on the selection part of EA. Since distance gives away information on the problem, it can also be leveraged in other parts of an EA to speed up the optimization even more. At the moment, there exist classical requirements to the design of crossover and mutation operators for an EA in a metric space \cite{droste2000metric, droste2003design, jansen2013evolutionary}. A practitioner, who wants to use metrics, needs to redesign the operators accordingly, which requires the application of methods from categories 1 and 2.

\textbf{Our contribution:} We propose two versions of the mutation operator which employs a given distance and allows switching it without the necessity to amend the algorithm. The first developed mutation operator uses information about possible values of the distance. The second operator does not require any knowledge of the structure of the distance and uses it as a black-box. The application of such mutation operators in an EA does not require any additional effort from a practitioner to switch one metric to another and has great benefits where a natural representation of the problem lacks strong causality. 


\section{Preliminaries}

\newcommand{\normd}{\mathcal{N}(\mathbf{0}, \mathbf{I})}

Evolutionary algorithms (EAs) are a class of search and optimization algorithms inspired by the principles of natural selection and Darwinian evolution. These algorithms are based on the idea that a population of solutions to a problem can evolve and improve over time through the application of selection, reproduction, and variation operators.
For example, the algorithm, (1+1)-ES, uses an iterative procedure that mutates the search point of a single individual $x \in \R^n$ by a multivariate normal distribution, $x' = x + \sigma \cdot \normd$, and assigns $x \leftarrow x'$ iff $f(x') \leq f(x)$. 

\update{During the optimization process an EA samples elements of the search space $X$ and evaluates the values of fitness function $f$ in the sampled points. While doing it, EA ideally balances exploration/exploitation. It explores the search space by sampling elements with significantly different structures, which can be seen as gaining knowledge of the  problem. Reversely, an EA exploits when structures of consequently sampled elements are similar. The significance of the structural difference between sampled solutions can be measured with metric $d$, which is usually considered as the distance between elements is $X$. Despite the fact that the optimization problem is completely specified when $X$ and $f$ are given, it is convenient to talk about the landscape of the problem when the exploration/exploitation capabilities of an EA are analyzed. The landscape of the optimization problem is classically defined as follows. }

\begin{definition}[Landscape \cite{reeves2000fitness}] For the problem $f$ defined on the metric space $X$ with metric $d$ the landscape is a triple $(X, f, d)$.
\label{def:landscape}
\end{definition}

The following algorithms are used in this work for the basis of our proposed mutation operators:

\emph{$(1+\lambda)~\text{EA}$} is an elitist evolutionary algorithm without crossover. Given the fitness function $f : X \to \R$, such algorithm maintains the most fit solution and iteratively samples modifications of this solution. Each modification is obtained using the given parameterized distribution $\mathcal{D}$ over the number of modified components \cite{doerr2016optimal}.

\emph{$(1+(\lambda,\lambda))~\text{EA}$} is an extension of $(1+\lambda)~\text{EA}$ which uses crossover with bias $c$. At first, this algorithm samples $\lambda$ solutions using mutation, then it picks the best mutant and recombines it $\lambda$ times with the best-so-far solution. Each recombination applies crossover in such a way that for every component the bit is taken from the first parent with probability $(1-c)$, and from the second parent with probability $c$. The best-so-far solution is updated with the fittest individual after such recombinations. This algorithm has proven itself well in theory \cite{antipov2019tight, antipov20201}.

\emph{Randomized Local Search} (RLS) is a stochastic local search algorithm. RLS iteratively explores the neighborhood of a current solution and selects a new solution based on a randomized criterion that takes into account the quality of the neighboring solutions. RLS is characterized by its simplicity, efficiency, and ease of implementation. In this work we apply an extension of this algorithm to integer search spaces, so-called RLS-ab \cite{doerr2018static}. This extension controls the set of values of size $n$, which identifies how quickly the corresponding component is updated. The control is done independently for every value. When the new mutant is better than the best-so-far solution, the value is increased, otherwise, it is decreased.

\emph{$(1+\lambda)~\text{EA}_\text{ab}$} This is our simple extension of RLS-ab. The update adds a possibility to create $\lambda$ individuals on every iteration and sample the number of modified components from the binomial distribution with success probability $1/n$. The operator to make mutation for every selected component stays the same, including the update rule of the parameters.

\emph{Univariate Marginal Distribution Algorithm} (UMDA)~\cite{larranaga2001estimation} is another population-based stochastic optimization algorithm. UMDA operates by sampling and updating the probability distribution of each variable in the solution vector independently, and generating new candidate solutions using a random sampling process based on the updated probability distributions. Every iteration it samples $\lambda$ individuals and uses $\mu$ best of them to update the probability distribution. This algorithm does not use the notions of the neighbourhood and so it has the potential to be efficiently applied even when SC distance does not exist for the problem. However, it is suggested to use a large enough size of $\lambda$ to solve problems with difficult properties \cite{doerr2020univariate}, so this algorithm may not be applicable in practical applications right away. We apply this algorithm to a relatively cheap function when we design our mutation operator. In this work, UMDA is adapted to handle integer domains. To achieve this we maintain the estimated probability of every integer in every component.


\section{High-level overview of the proposed approaches}

Given the definition \ref{def:landscape} of a landscape, we say that an algorithm becomes landscape aware if it uses the specific distance $d$ for solving the optimization problem apart from $X$ and $f$ that are always used by RSH. In the scope of this work, we only propose ways to use this distance in the mutation operator. We will call a mutation operator \textit{distance-driven} if it uses such distance to generate mutants. Accordingly, we will say that an algorithm is distance-driven if it uses a distance-driven mutation operator. 

Let us consider RSH $\mathcal{A}$ which applies a mutation operator when it solves an optimization problem in the landscape $(X, f, d')$. We make an assumption that $d'$ is not a SC distance for $f$. At the same time, we assume that SC distance $d \ne d'$ for $f$ is given to us by an oracle. In this section we propose a way to change $\mathcal{A}$ to a distance-driven algorithm dd-$\mathcal{A}$ which solves the optimization problem (at least partially) in the landscape $(X, f, d)$. The optimization is performed in the new landscape completely when $\mathcal{A}$ uses only mutations to sample new solutions. However, we leave distance-driven crossover operators for future work, so if $\mathcal{A}$ uses recombination in addition to mutation, then dd-$\mathcal{A}$ partially works in the old landscape.

Both our distance-driven mutation operators rely on a parameterized family of distributions over all possible step sizes. In Section \ref{sec:dd-mutation-known-structure}  we explain what we exactly mean by this. In Section \ref{sec:dd-mutation-known-structure} we assume that the distribution is defined by hand, which is possible when the so-called structure of the distance is known. Then in Section \ref{sec:dd-mutation-black-box}, we show how to create this distribution automatically for any given distance. The distance-driven mutation operator itself is given in Algorithm \ref{alg:mutation} and it is applied in both proposed methods, but with different distributions. Consequently, given SC distance $d$ and distribution $\mathcal{D}$ defined in Section \ref{sec:dd-mutation-known-structure} we transform $\mathcal{A}$ to dd-$\mathcal{A}$ by substituting distance-driven mutation operator to $\mathcal{A}$ instead of the usual mutation operator.

\section{Proposed method for the distance with known structure} \label{sec:dd-mutation-known-structure}

Traditionally, mutation creates a structural change in the individual. The change is usually defined by the probabilistic distribution over different structural changes. The shape of the distribution reflects the intuition about the search. For example, in local search methods, the distribution is defined over only small mutations, that likely do not lead to quitting the basin of attraction. In global search methods, such as Evolutionary Strategies (ES), the distribution is defined over all the search space, but smaller changes happen with higher probability. Even though the mutation itself is sampled from this distribution, the significance of this change is defined w.r.t. the distance metric. Classically, it is assumed that the distribution is equipped with such a distance metric, that considers components independently and a small change in the value of every component contributes a small addition to the total value of distance. When the metric is different, but the same algorithm is applied, the mutations sampled from the same distribution likely do not give the properties that are expected from the perturbations.  

Let us consider mutation $\mu : X \to X$ that happens at some point in the optimization algorithm. Note that $\mu$ and $\mu'$ as mutations at different time points may be different due to the stochasticity of the heuristic search. We will say that \textit{step size} of mutation w.r.t. distance $d$ is the value $s \coloneqq d(x, \mu(x))$.
The probabilistic distribution that is used in the mutation operator defines the step size of the operator. This step size may be sampled from the distribution explicitly, or implicitly, as happens in ES when the mutation itself is sampled. In the mutation operator that we design, we do not use anything from the distribution except the information on the step size. Hence, we consider the distribution defined over step sizes. 

\update{Let us define what we denote as the distribution over step sizes more precisely. First of all, we denote the set of all pairwise distances between points in the search space $S = \{d(x,y) \mid x, y \in X\}$ as \textit{structure of the distance}. In this section, we assume that $S$ is known to the designer of the algorithm. In order to allow balancing exploration/exploitation in the algorithm we add parametrization to the distribution, which significantly improves the performance of EA \cite{eiben1999parameter, antonov2021blending}. All the parameters of the distribution are denoted with $\alpha$. This vector controls the shape of the distribution and is updated outside of the mutation operator according to the rules defined in the distance-driven algorithm. Given parameters $\alpha$, individual $x \in X$ which is being perturbed and metric $d$, the distribution over step sizes from $x$ is: }
\begin{equation}\mathcal{D}_{x,\alpha} \coloneqq \{\Pr_\alpha\left[d(x, \mu(x)) = d(x, y)\right] \mid y \in X\} \label{eq:distribution-x}\end{equation}

\update{For the given $\alpha$, distance-driven algorithm possess the following set of distributions in order to apply distance-driven mutation operator at any point of $X$: }
\begin{equation}\mathcal{D}_{\alpha} \coloneqq \{\mathcal{D}_{x,\alpha} \mid x \in X\} \label{eq:distribution-alpha}\end{equation}

\update{Then the parametrized family of sets of distributions is given by:}
\begin{equation}\mathcal{D} \coloneqq \{\mathcal{D_{\alpha}} \mid \alpha\} \label{eq:distribution}\end{equation}

We will say that $\mathcal{D}$ defined in Eq. \ref{eq:distribution} is \textit{distributions family}. The choice of this distribution family is left to the user. To choose $\mathcal{D}$ it is sufficient to define $\mathcal{D_{\alpha}}$ in Eq. \ref{eq:distribution-alpha} and so it is sufficient to define $\mathcal{D}_{x,\alpha}$ in Eq. \ref{eq:distribution-x}. Therefore it is sufficient to define the parametrized probability of mutation from any element in $X$ to any element in $X$. In order to do this, it is necessary and sufficient to know the set $S$. 

Our distance-driven mutation operator generates another individual that steps away from the initial one to the distance closest to the given step size. To implement this generation we solve another optimization problem as written in Algorithm \ref{alg:mutation}.

\begin{algorithm}
    \caption{Distance-driven mutation operator of individual $x \in X$ given distance $d : X^2 \to \R$, distributions family $\mathcal{D}$ and distribution parameters $\alpha$}
    \begin{algorithmic}[1]
        \State{$\mathcal{D_{\alpha}} \in \mathcal{D}$}
        \Comment{Choose parametrized distribution from $\mathcal{D}$}
        \State{$\mathcal{D}_{x,\alpha} \in \mathcal{D}_{\alpha}$}
        \Comment{Choose distribution for the given point $x$}
        \State{$s \sim \mathcal{D}_{x, \alpha}$}
        \Comment{Sample mutation strength}
        \State{$y \gets \argmin_y\{\left|d(x,y)-s\right|\}$} \label{alg:mutation:opt}
        \Comment{Internal optimization}
        \State{\Return{$y$}}
    \end{algorithmic}
    \label{alg:mutation}
\end{algorithm}

In line \ref{alg:mutation:opt} of Algorithm \ref{alg:mutation} the optimization problem is solved. The properties of the function optimized in this line are mostly defined by the properties of distance $d$. Note that Hamming distance (or Euclidean distance) $h$ is not SC distance for this function, because otherwise $h(x, y) < \eps \implies d(x, y) < \eps \implies \left|f(x) - f(y)\right| < \delta$ which contradicts our assumption that $h$ is not SC distance for fitness function $f$. In order to optimize this function, we apply UMDA with a large $\lambda$ and a big budget.

Application of this version of the mutation operator requires the user to know the structure of distance and define the distributions family. Obtaining knowledge of the distance structure may require expertise in the domain of the problem. Moreover, the size of $S$ may be extremely big, so defining the distributions family may not be feasible in some cases. To free the user from the necessity to learn this structure, we propose the following extension of the mutation operator, which automatically does both: explores the structure of distance and creates distributions family.



\section{Proposed method for black-box distance} \label{sec:dd-mutation-black-box}

The main purpose of this section is to propose a method that is able to extend the perturbation operator for the distance with an unknown structure. This allows the decoupling of an expert in optimization and an expert in the field of the domain. We expect that the distance created by an expert in the domain is well-defined. It means that the distance is SC and on top of that we also expect certain structural properties, which we summarize in the following assumption. 

\begin{assumption}
    For any points $x, y \in X$:
    \begin{enumerate}
        \item Variability: $\left|\{d(x, z) \mid z \in X\}\right| \gg 1$;
        \item Consistency: $\min\{d(x,z) \mid z \in X \setminus \{x\}\} \approx \min\{d(y,z) \mid z \in X \setminus \{y\}\}$; moreover $\max\{d(x,z) \mid z \in X\} \gg \min\{d(x,z) \mid z \in X \setminus \{x\}\}$;
        \item Cheapness: wall-clock time required for computation of $d(x,y)$ is significantly smaller than the wall-clock time required for computation of the objective function at any point.
    \end{enumerate}
\end{assumption}

The most important assumption for our methods is cheapness because the inner optimization problem is solved for every application of the mutation operator, which is normally called as often as the objective function. When the metric does not satisfy this property our method becomes very time-consuming. If the variability assumption is not upheld for many points, then the practitioner, who knows this fact, has already enough knowledge to use the previous method and create distributions family $\mathcal{D}$ based on their tastes. The consistency assumption is needed to save computational resources when we explore the space of possible values of the given metric. Our method is still applicable if this condition is not upheld for a subset of the search space, but for every point in this subset, we need to run a new exploration procedure, which becomes time-consuming. 

In the case of black-box distance, it is impossible to define a distributions family $\mathcal{D}$ in advance. Here we propose a method to create such distributions family that can be used in Algorithm \ref{alg:mutation}.
Designing this distributions family for the general case of the metric is challenging because of the two following problems. 
1) \textit{Different scales:} For different search points $x$, $y$ and different metrics $d_1$, $d_2$ it may be the case that $d_1(x,y)/d_2(x,y) \gg 1$. 
2) \textit{Different gaps:} For different search points $x,y \in X$ and different metrics $d_1, d_2$, the values $v_i \coloneqq \left|\{z \in X \mid \max\{d_i(x, z), d_i(y, z)\} < d_i(x, y)\}\right|$ may be significantly different for $i \in \{1, 2\}$. 

At first, we propose a general transformation of the distance metric to solve both those problems. Then we define the distributions family over the transformed values of distance. This transformed distance and distributions family are used in Algorithm~\ref{alg:mutation} in the same way as in the previous subsection.

\subsection{Transformation of the distance}

Given metric $d$, for every point $x \in X$ we map $A_{d, x} = \{d(x, y) \mid y \in X\}$ to the spread subset of $(0, 1)$ with a monotonic function \begin{equation}\tau_{\gamma}(x, y) = 1 - \exp{(-\gamma^2d^2(x,y))} \label{eq:mapping} \end{equation} The second term in the mapping is the well-known Gaussian function. Let us define the set $\mathcal{I}_{\gamma} = \{\tau_{\gamma}(x, y) \mid y \in X \setminus \{x\}\}$. Such mapping transforms $A_{d, x}$ to $\mathcal{I}_{\gamma} \subset (0,1)$ for any metric $d$ and so allows to have the same ranges of distances, which solves the first problem we mentioned. The parameter $\gamma$ of Eq. \ref{eq:mapping} is chosen in such a way that $\mathcal{I}_{\gamma}$ is as spread as possible. This requirement ensures that all the values of $\mathcal{I}_{\gamma}$ do not collapse to a very dense segment where different values of distance are not distinguishable given the limitations of the representation of floating point numbers.

To satisfy this requirement we consider the following problem regarding $\eps_1, \eps_2$:

\begin{align}
    \min \quad & \eps_1 \\
    \min \quad & \eps_2 \\
    \textrm{s.t.} \quad & \min{\mathcal{I}_{\gamma}} \le \eps_1 \label{constr:eps1}\\ 
    & \max{\mathcal{I}_{\gamma}} \ge 1 - \eps_2 \label{constr:eps2}   
\end{align}

The constraint \ref{constr:eps1} defines the upper bound on the smallest mapped distance, analogously constraint \ref{constr:eps2} defines the lower bound for the largest mapped distance. For the purposes of our application, we simplify the problem. We aim to find sufficiently small $\eps_1$ and ensure that $\eps_2$ is small as well. Specifically, we iterate over values of $\eps_1$ in the range $(10^{-4}, 10^{-2})$ with step $10^{-4}$ and stop when we find smallest $\eps_1 \ge \eps_2$ such that $\eps_1, \eps_2$ satisfy the constraints \ref{constr:eps1}, \ref{constr:eps2}. 

To check if the given value of $\eps_1$ satisfies the constraints we simplify them as follows. Using monotonic property of Eq.~\ref{eq:mapping} we have: 
\begin{equation*}
    \begin{cases}
        1 - \exp{(-\gamma^2 (\min{A_{d,x}})^2)} \le \eps_1 \\
        1 - \exp{(-\gamma^2(\max{A_{d,x})^2)}} \ge 1 - \eps_2        
    \end{cases}
\end{equation*}

which is transformed to:
\[\dfrac{\sqrt{-\ln(\eps_2)}}{(\max{A_{d,x}})^2} \le \gamma \le \dfrac{\sqrt{-\ln(1-\eps_1)}}{(\min{A_{d,x}})^2}\]

We combine this with the additional condition $\eps_1 \ge \eps_2$ and obtain the following:
\[(1-\eps_1)^{\left(\dfrac{\max{A_{d,x}}}{\min{A_{d,x}}}\right)^2} \le \eps_1\]

To estimate the values $\zeta_{\text{max}} \coloneqq \max{A_{d,x}}$ and $\zeta_{\text{min}} \coloneqq \min{A_{d,x}}$ we consider fixed values $t, T \in \R : 0 < t < 1 < T$ and such sequence $(z_i)_{i = -k, \dots, k} $ that:
\begin{equation*}
    \begin{cases}
        z_i \sim \text{u.a.r.}(X \setminus \{x\}), & \text{if } i = 0 \\ 
        d(x, z_i) \geq Td(x, z_{i-1}), & i > 0 \\
        d(x, z_i) \leq td(x, z_{i+1}), & i < 0
    \end{cases}
\end{equation*}

As approximation of $\zeta_{\text{max}}$ and $\zeta_{\text{min}}$ we take $\widehat{\zeta_{\text{max}}} \coloneqq d(x, z_k)$, $\widehat{\zeta_{\text{min}}} \coloneqq d(x, z_{-k})$ accordingly. Steps above to compute $\eps_1, \eps_2, \gamma$ are summarised in Algorithm~\ref{alg:gamma}.

\begin{algorithm}
    \caption{Computation of the parameters $\eps_1, \eps_2, \gamma$ for the given point $x \in X$}
    \begin{algorithmic}[1]
        \State{$z_0 \gets \text{u.a.r.}(X \setminus \{x\})$}
        \For{$i \gets 1, \dots, k$}
            \State{$z_i \gets z \in \left\{z \in X \mid d(x, z) \geq T\cdot d(x, z_{i-1})\right\}$} \label{alg:gamma:zmax}
            \Comment{Optimize}
            \State{$z_{-i} \gets z \in \left\{z \in X \setminus \{x\} \mid d(x, z) \leq t\cdot d(x, z_{i+1})\right\}$} \label{alg:gamma:zmin}
            \Comment{Optim.}
        \EndFor
        \State{$\widehat{\zeta_{\text{max}}} \gets d(x,z_k)$}
        \State{$\widehat{\zeta_{\text{min}}} \gets d(x,z_{-k})$}
        \For{$j \gets 1, \dots, c$}
            \State{$\eps_1 \gets 10^{-4}j$}
            \State{$\eps_2 \gets \exp{\left[\left(\dfrac{\widehat{\zeta_{\text{max}}}}{\widehat{\zeta_{\text{min}}}}\right)^2\ln{(1-\eps_1)}\right]}$} \label{alg:gamma:eps2}
            \State{\textbf{if} $\eps_2 \le \eps_1$ \textbf{then break end if}}
        \EndFor
        \State{$\gamma \gets \dfrac{1}{2}\left(\dfrac{\sqrt{-\ln(\eps_2)}}{\widehat{\zeta_{\text{max}}}^2} + \dfrac{\sqrt{-\ln(1-\eps_1)}}{\widehat{\zeta_{\text{min}}}^2}\right)$}
        \State{\Return{\{$\eps_1, \eps_2, \gamma$\}}}
    \end{algorithmic}
    \label{alg:gamma}
\end{algorithm}

Algorithm \ref{alg:gamma} takes as input a point from $X$. When a certain point $p \in X$ is passed as input to Algorithm \ref{alg:gamma} we say that the output values $\eps_1, \eps_2, \gamma$ are \textit{computed for} the point $p$. Following our consistency assumption, the values of $\eps_1$ and $\eps_1'$ are similar when computed for different points $x, x' \in X$.
Using the same assumption we have: $d(x, z_{k}) \gg d(x, z_{-k})$. When the optimization problems in lines \ref{alg:gamma:zmax}, \ref{alg:gamma:zmin} of Algorithm \ref{alg:gamma} are solved precisely we obtain: $\widehat{\zeta_{\text{max}}} \gg \widehat{\zeta_{\text{min}}}$. So we neglect the small change in $\left(\frac{\widehat{\zeta_{\text{max}}}}{\widehat{\zeta_{\text{min}}}}\right)^2$ and argue that values $\eps_2$ and $\eps_2'$ are very close, when computed as in line \ref{alg:gamma:eps2} of Algorithm \ref{alg:gamma} for different points $x, x' \in X$. We take advantage of this and compute $\eps_1, \eps_2, \gamma$ only once for a point $x \in X$ chosen uniformly at random. 

Following the variability assumption, $\left|A_{d,x}\right| \gg 1$ which makes elements of $\mathcal{I}_{\gamma} = \{\tau_{\gamma}(x,y) \mid y \in X\}$ located very close to each other in $\mathcal{I}_{\gamma}$. It means that there are no significant gaps between values in  $\mathcal{I}_{\gamma}$, which solves the second problem. 

In our implementation, we used the following constants: $k = 10, T = 2, t = 1.2, c = 100$. To find an element from the set defined in line \ref{alg:gamma:zmax} we apply UMDA to maximize the objective function $f_1(z) \coloneqq d(x, z) - T\cdot d(x, z_{i-1})$ with stopping criteria that, apart from limiting the number of function evaluations, has condition $f_1(z) \ge 0$. Similarly, to compute $z_{-i}$ in line \ref{alg:gamma:zmin} we apply UMDA with the objective function $f_2(z) \coloneqq t\cdot d(x, z_{i+1}) - d(x, z)$ and stopping criteria that includes conditions $f_2(z) \ge 0 $ and $f_2(z) < t\cdot d(x, z_{i+1})$. The second additional condition is needed to ensure that $z_{-i} \ne x$.

\subsection{Distribution over the transformed values}

In this subsection, we derive distribution $\mathcal{D}_{x,\alpha}$ for arbitrary element $x \in X$ and parameters $\alpha$. Since the values $\eps_1, \eps_2, \gamma$ are the same for all the points, the distributions $\mathcal{D}_{x,\alpha}$ are the same for all the points in $X$ when parameters $\alpha$ are fixed. To be more concise, in this section, we will denote $\mathcal{D}_{x,\alpha}$ as $\mathcal{D}$. 

The distribution over $\mathcal{I}_{\gamma}$ is chosen to be continuous to avoid computation of the exact value $\left|\mathcal{I}_{\gamma}\right|$. The fact that $\mathcal{I}_{\gamma}$ has finite size does not corrupt the algorithm, because of the big density of points in $\mathcal{I}_{\gamma}$. Moreover, the optimization problem in Algorithm \ref{alg:mutation} may not be solved precisely even when the distribution is discrete. It means that for a random variable $\xi \sim \mathcal{D}$ the value $\xi - \min\{\left|d(x,y)-\xi\right|\}$ may not be zero in practice, but close to zero in the case of discrete and continuous distributions. Then, the distribution is chosen as such $\mathcal{D}^*$ that have maximal differential entropy over continuous distributions with fixed mean $m \in \mathcal{I}_{\gamma}$: \begin{equation} \mathcal{D}^* = \argmax_{\mathcal{D}}\{H(\mathcal{D}) \mid \E(\xi \sim \mathcal{D}) = m\} \label{eq:cont-distribution} \end{equation}

The maximal entropy is enforced to obtain an unbiased mutation operator \cite{rudolph2012handbook}. We fix the mean of this distribution to allow controlling the step size during the optimization process. Distributions with fixed moments of higher order and other properties are left for future works. This is done because enforcing other properties will make it much more complex to solve the optimization problem defined in Eq. \ref{eq:cont-distribution}. On the other hand, distribution with a fixed mean and maximal entropy is already enough to efficiently solve problems with different complex properties, when the algorithm adjusts the mean during the optimization process. For example, one of $(1+\lambda)$ EA with mutation rate control optimizes a number of difficult pseudo-boolean functions faster than competitors \cite{doerr2019benchmarking}. The step size of mutation in $(1+\lambda)$ EA is sampled from the Binomial distribution. Mutation rate control is equivalent to controlling only the mean of this distribution. At the same time, Binomail distribution is the maximal entropy distribution among discrete distributions with the fixed mean and fixed support \cite{harremoes2001binomial}. This motivates the hope, that our mutation operator with distribution from Eq. \ref{eq:cont-distribution} has the potential to assist optimization better than mutation operators with other distributions, given that a smart enough method to control the mean is used. 

In order to find the probability density function (pdf) $\mathcal{P}$ of distribution defined in Eq. \ref{eq:cont-distribution} we consider the following optimization problem:

\begin{align}
    \argmax\limits_{\mathcal{P}} \quad & {-\int\limits_{\eps_1}^{1-\eps_2}{\mathcal{P}(x)\ln\mathcal{P}(x)dx}} \label{entropy} \\ 
    \textrm{s.t.} \quad & \int\limits_{\eps_1}^{1-\eps_2}{\mathcal{P}(x)dx} = 1 \label{constraint1} \\ 
    & \int\limits_{\eps_1}^{1-\eps_2}{\mathcal{P}(x)xdx} = m \label{constraint2}
\end{align}


The following solution of Eq. \ref{entropy} is unique for such $\lambda_0, \lambda_1 \in \R$ that satisfy the constraints Eq. \ref{constraint1}, Eq. \ref{constraint2} \cite{jaynes1957information, jaynes1957information2, cover1999elements}: \begin{equation} \mathcal{P}(x) = \exp{(\lambda_0 + \lambda_1x)} \label{eq:general-solution} \end{equation}

Substitution of Eq. \ref{eq:general-solution} to Eq. \ref{constraint1}, Eq. \ref{constraint2} and integration, gives us the following system:

\begin{equation*}
    \begin{cases}
        \dfrac{\exp{(\lambda_0)}}{\lambda_1}\left(e^{\lambda_1(1-\eps_2)} - e^{\lambda_1\eps_1}\right) = 1 \\
        \dfrac{\exp{(\lambda_0)}}{\lambda_1^2}\left(e^{\lambda_1(1-\eps_2)}(\lambda_1(1-\eps_2)-1) - e^{\lambda_1\eps_1}(\lambda_1\eps_1-1)\right) = m
    \end{cases}
\end{equation*}

Combining those two equations into one with condition that $\lambda_1 \ne 0$ gives us the following:

\begin{multline}    
    e^{\lambda_1(1-\eps_2)}(\lambda_1(1-\eps_2)-1) - e^{\lambda_1\eps_1}(\lambda_1\eps_1-1) = \\ \lambda_1 m \left(e^{\lambda_1(1-\eps_2)}-e^{\lambda_1\eps_1}\right)
    \label{eq:leftright}
\end{multline}

If the solution of Eq. \ref{eq:leftright} exists regarding $\lambda_1$ then it is unique because the solution of Eq. \ref{entropy} is unique. The left part of Eq. \ref{eq:leftright} converges to 0 for $\lambda_1 \to -\infty$ and converges to 0 for $\lambda_1 \to 0$. The right part of Eq. \ref{eq:leftright} is infinitely big for $\lambda_1 \to -\infty$ and converges to 0 for $\lambda_1 \to 0$. Moreover for $\lambda_1 \to -0$ the left part is greater than the right part. Since the functions of $\lambda_1 < 0$ in both parts are continuous there exists a solution of Eq. \ref{eq:leftright} for $\lambda_1 < 0$. To find the solution we apply binary search with bounds $(-K, 0)$, where $K$ is sufficiently big.



\begin{figure*}[!h]
    \includegraphics[width=\textwidth]{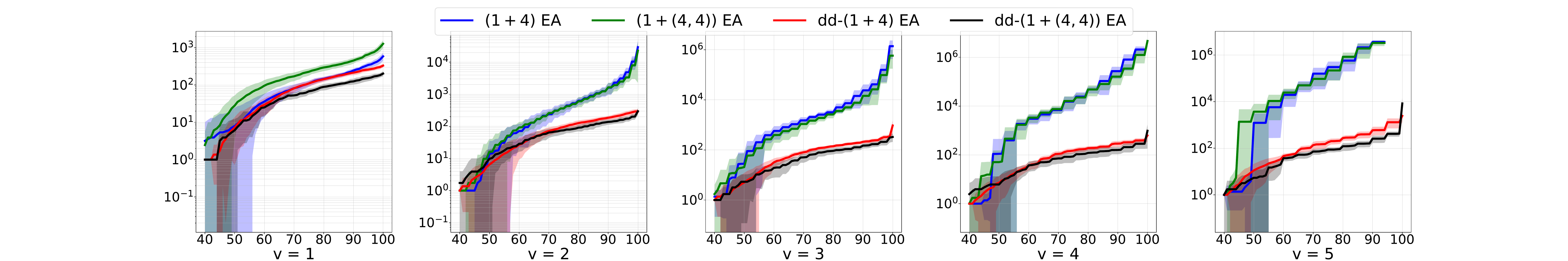}
    
    \caption{Average parallel optimization time and standard deviation of the considered algorithms (x-axis: best-so-far value of fitness function; y-axis: average number of iterations) on the \Rugg with $v \in [1, 5]$. Dimensionality $n = 100$, $11$ runs for every algorithm. UMDA is used as an internal optimizer of distance-driven algorithms with parameters $\mu = 50, \lambda = 100$, budget = $1000$. The limit on the number of iterations is $5\cdot10^6$.\\}
    \label{fig:onemax}
\end{figure*}

\begin{figure*}[!h]
    \includegraphics[width=1\textwidth,trim=38mm 24mm 37mm 2mm,clip]{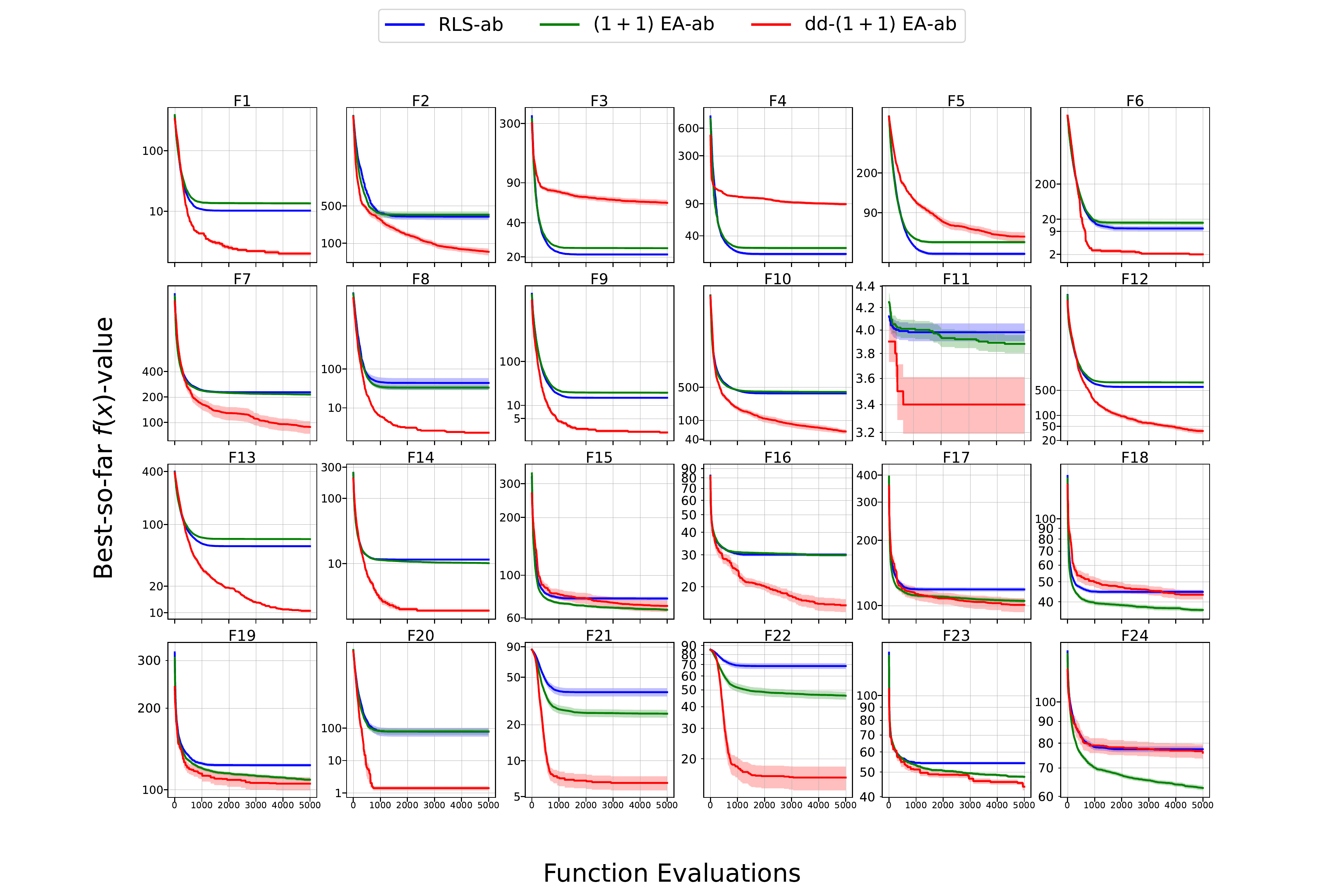}

    \caption{Convergence of the considered algorithms (x-axis: number of fitness evaluations; y-axis: best-so-far value of regret and its standard error) on the transformed \bbobmixint suite. Dimensionality $n = 40$, cardinality of every component is $100$, $100$ runs for RLS-ab and $(1+1)$ EA-ab, $10$ run for dd$(1+1)$ EA-ab. UMDA is used as an internal optimizer of the distance-driven algorithm with parameters $\mu = 10^2, \lambda = 10^3$, budget = $4\cdot 10^4$. The limit on the number of fitness function evaluations is $5000$.}
    \label{fig:bbob}
\end{figure*}

\section{Experiments}
In this section we outline the different experimental setups used to validate the proposed contributions. This is done using two experiments: binary \OneMax-based problems to show the effectiveness of using distance-driven permutation in an easy-to-analyze use-case, and a discretized version of the BBOB benchmark suite to show the generalizability of the proposed approach.

\subsection{Experimental Setup}

\paragraph{Binary problems.}
In order to benchmark Algorithm~\ref{alg:mutation} as it is, we consider a binary \OneMax-based problem in $n$ dimensions (i.e. with search space $X = \{0,1\}^n$). We adopt \Rugg problem based on \OneMax and implement it as in \textit{W-Model} suite \cite{weise2018difficult}. 
In our experiments, we define \OneMax as $l_1(x) \coloneqq \left|x\right|$ ($L_1$-norm of binary string $x$). 
\Rugg uses a given permutation $\pi$ and applies it to the objective values of the underlying function, which is \OneMax in our case. We define such \Rugg as: $r_{\pi}(x) \coloneqq \pi^{-1}(l_1(x))$. We define distance for this problem as $d(x,y) \coloneqq \left|r_{\pi}(x) - r_{\pi}(y)\right|$. Note that such distance satisfies all the requirements of the distance metric because it is based on the absolute value. Moreover, this is SC distance. The permutation is stored implicitly, such that for every integer $0 < v < n$:

\begin{equation*}
    \begin{cases}
        r(x) = l_1(x) + v - 2(l_1(x) \bmod v) - 1, & \text{if } \lfloor l_1/v \rfloor \ne \lfloor n/v \rfloor \\
        r(x) = l_1(x) + v_1 - 2k - 1, & \text{if } \lfloor l_1/v \rfloor = \lfloor n/v \rfloor
    \end{cases}
\end{equation*}

Here $v_1 = (n \bmod v)$ and $k = (l_1(x) \bmod v_1) - ((\lfloor n/v \rfloor v)\bmod v)$. In our implementation, we considered $v \in [1, 5]$. Note that with $v = 1$ \Rugg is \OneMax. Greater values of $v$ create dissipative intervals of size $v$, for example, for $v = 5$ the objective value 99 of \OneMax is permuted to 95, 99 to 96, \dots, 95 to 99.

We extended both $(1+\lambda)$ EA and $(1+(\lambda, \lambda))$ EA with our distance driven mutation operator. The modified algorithms are called dd-$(1+\lambda)$ EA and dd-$(1+(\lambda, \lambda))$ EA accordingly. Both algorithms are considered without any parameter control. In the implementation of Algorithm \ref{alg:mutation} we implemented Binomial distribution with $n$ Bernoulli experiments with success probability $1/n$. Note that the value sampled from this distribution may be larger than the maximal distance reachable from point $x$ (for example when it is of the form $11..100..0$). In this case, the optimization in line \ref{alg:mutation:opt} of Algorithm \ref{alg:mutation} is terminated when it runs out of budget. The parameters of UMDA taken in the experiments on binary problems are the following: $\mu = 50, \lambda = 100$, budget = $1000$.

\paragraph{Integer problems.}
We consider integer problems to demonstrate the advantages of the distance-driven mutation operator when the distance satisfies the assumptions. The domain of those problems is $X = \Pi_{i = 1}^n \{0, c_i - 1\}$,  where $(c_i)_{i=1\dots n}$ are cardinalities. The problems are taken from \bbobmixint suit \cite{tuvsar2019mixed}. We extended the problems to get rid of continuous variables and support the arbitrary cardinality of every component, the rest details on these problems were left unchanged. To show the advantages of our mutation operator, we transform those problems in our experiments. For the given permutation $\pi$ and function $f$ from BBOB-MIXINT we define the transformed function $Tf(x) \coloneqq f\left[(\pi^{-1}(x_i))_{i = 1, \dots, n}\right]$. Here $x_i$ stands for the $i$-th component of $x$. Note that here permutation is the same for all the dimensions. We implemented this transformation because in our experiments the cardinalities of all the dimensions are equal. When permutation is applied to the search space the neighbourhood structures are destroyed in every component, so Euclidean distance becomes ill-conditioned. We define distance for the transformed problem as $d(x, y) \coloneqq l_2(\left[(\pi^{-1}(x_i))_{i = 1, \dots, n}\right] - \left[(\pi^{-1}(y_i))_{i = 1, \dots, n}\right])$. This distance takes inverse permutations of every component and computes $L_2$-norm of the difference of obtained vectors. In all of the experiments, the same permutation was used for search space transformation.

In the experiments we consider $n = 40$ dimensions, each one with cardinality $c = 100$ for every problem in the suit \bbobmixint. We extended UMDA to the domains where the cardinality of every component is greater than 2. The parameters of this extended UMDA taken in the experiments on those problems are the following: $\mu = 10^2, \lambda = 10^3$, budget = $4\cdot 10^4$.

In our experiments we maximized negative regret, meaning the value $f(x^*) - f(x)$, where $f$ is a problem, $x^*$ is the global optimum of $f$, $x$ is a candidate solution. At the same time, we plot positive regret in Figure \ref{fig:bbob} to avoid confusion.

We extended RLS algorithm with our mutation operator. Note that the new algorithm does not do local search anymore because the distribution is defined over all found values of distances. We control the mean of the distribution in the following way. Assume that the mutation operator produced individual $y$, and the best-so-far solution is $x$. Then if $f(y) > f(x)$ we increase the mean (which increases the expected step size) $m = a\cdot m$. If $f(y) < f(x)$ then we reduce the mean $m = b\cdot m$. The rest of the steps (except mutation and parameter update) are the same as in RLS algorithm. We call this modified algorithm dd-$(1+1)~\text{EA}_{\text{ab}}$. In the implementation, we take values $a = 1.001, b = 0.999$. To simplify the numerical solution of Eq. \ref{eq:leftright} we took $\eps_1 = 0, \eps_2 = 1$ at that equation (and only there). In this case the solution regarding $\lambda_1$ always belongs to $[-1/m, 0)$.

The implementation of all considered algorithms, problems and experiments can be accessed at Zenodo \cite{zenodo_2023_7897290}.

\subsection{Results}

\paragraph{Binary problems.} In Figure \ref{fig:onemax} we compare the performance of the described algorithms on \Rugg problem with different values of $v$. Algorithms with modified mutation operators performed much better than their analogues in all of the considered cases of $v > 1$. For $v = 1$ the performance of the algorithm is almost the same, indeed distance-driven mutation operator uses the same step sizes because the distribution is the same for all the algorithms. However, $\text{dd}-(1+(4,4))~\text{EA}$ slightly outperforms dd-$(1+4)$ EA on most of the points, except the final points for $v = 4$ and $v = 5$. When an individual is in the local optima, the mutation-only algorithm waits in this optima until a sufficiently big step size is sampled from the distribution. At the same time, crossover adds additional variability which helps to escape the local optima. This is the reason for the slight advantage of $\text{dd}-(1+(4,4))~\text{EA}$ on the most of optimization trajectory. At the same time, the performance of both algorithms is significantly better than their analogues and the gap between them grows with the growth of $v$.

\paragraph{Integer problems.} In Figure \ref{fig:bbob} we compare the performance of dd-$(1+1)~\text{EA}_{\text{ab}}$ against two algorithms that are not aware of SC distance for the problems. As expected, for the majority of functions, specifically for F1, F2, F6-F14, F16, and F20-F22 the increase in performance of dd-$(1+1)~\text{EA}_{\text{ab}}$ is significant compared to the other algorithms. It shows that our algorithm manages to explore the search space well enough to find the values of parameters that allow the mutation operator to generate small distances as well as big ones. However, we observe that performance is much worse on functions F3 and F4. For function F4, this may happen because both competitor algorithms got lucky with a permutation of the search space, which allowed them to jump to the local optima of better quality at the beginning, where they converged to the basin of attraction and got stuck. For function F3, the distance-aware algorithm may lose because it treats all the directions evenly. When it finds the ridge of F3, it continues to sample solutions outstanding to the sampled step size $s$. Vectors of all possible perturbations make a hyper ball of radius $s$ which intersects many other points apart from the points on the ridge. So the probability to improve the value of the objective function becomes very small. This is the reason why we observe a very slow convergence of distance-driven algorithms on this function. For function F24 distance-driven algorithm also loses to the competitor that is able to make jumps to bigger step sizes in the search space. The reason for this might be the fact that basins of attraction on F24 are small, so the algorithm needs to make small steps to not jump from one valley to another. This is the weak spot of the distance-driven algorithm because in order to find smaller values of distance it needs to spend much more time on internal optimization, which makes the algorithm significantly slower. On top of that, the algorithm stops making jumps to other basins of attraction, because, after the initial convergence, the mean of the distribution becomes too small. It happens because the parameter control method that we pick for dd-$(1+1)~\text{EA}_{\text{ab}}$ is rather simple. On the rest of the functions, the distance-driven algorithm slightly wins or shows approximately the same performance as competitors.

\section{Conclusion}

In this work, we propose two extensions of the perturbation operator. Both operators are applicable to a broader class of algorithms than just EAs. They can extend other RSHs that use the notion of distance to make changes in the solution, for example, Simulated-Annealing. The first extension relies on the structure of the distance. The user of this approach is expected to design the distribution over step sizes that they target to see in the perturbation operator. When the step size is sampled from the distribution, it is used to solve the inner optimization problem, where the mutant is generated. The goal of the inner optimization is to generate such an individual that is a number of steps away from the given individual to a distance close to the sampled step size. 

The second mutation operator uses distance as a black-box. It explores the search space prior to defining the distribution. The information about the problem that was obtained during the exploration is then used to define the distribution over the step size with maximal entropy and fixed value of the mean. To estimate the parameters of this distribution we solve the numerical optimization problem and repeat this optimization every time the mean changes. This allows us to control the strength of changes that are made in the mutation operator.

We observed the performance of the first operator on pseudo-boolean problems and the performance of the second operator on integer problems to empirically evaluate the proposed operators. From the results, distance-driven operators were found to do better on most functions. It shows the advantage of being aware of the distance during the optimization. However, the performance on some functions is worse, because of imperfections of parameter control, specificities of the considered problem, and due to the fact that our operator treats all the directions evenly.

In future work, we plan to identify directions in the metric space to assist the mutations to proceed in the most promising direction with higher probability. This should improve the performance of the functions F3. Moreover, it will allow us to apply the most advanced methods of parameter control that exist in Evolutionary Strategies in our mutation operators.

\begin{acks}
We would like to thank our collaborators at SRON for the real-world motivation that inspired this work.
\end{acks}

\bibliographystyle{ACM-Reference-Format}
\bibliography{bibliography}

\end{document}